\documentclass[10pt,twocolumn,letterpaper]{article}

\usepackage{cvpr}
\usepackage{times}
\usepackage{epsfig}
\usepackage{graphicx}
\usepackage{amsmath}
\DeclareMathOperator*{\argmin}{argmin}
\usepackage{amssymb}
\usepackage[title]{appendix}
\usepackage{graphbox}

\usepackage[breaklinks=true,bookmarks=false]{hyperref}
\cvprfinalcopy 

\def\cvprPaperID{1418} 

\begin{document}

\title{Semantic Graph Convolutional Networks for 3D Human Pose Regression}

\author{Long Zhao$^1$ \quad Xi Peng$^2$ \quad Yu Tian$^1$ \quad Mubbasir Kapadia$^1$ \quad Dimitris N. Metaxas$^1$\\
$^1$Rutgers University \quad $^2$Binghamton University\\
{\tt\small \{lz311,yt219,mk1353,dnm\}@cs.rutgers.edu, xpeng@binghamton.edu}
}

\newcommand{\p}[1]{\textbf{#1}}

\maketitle

\begin{abstract}
   In this paper, we study the problem of learning Graph Convolutional Networks (GCNs) for regression. Current architectures of GCNs are limited to the small receptive field of convolution filters and shared transformation matrix for each node. To address these limitations, we propose Semantic Graph Convolutional Networks (SemGCN), a novel neural network architecture that operates on regression tasks with graph-structured data. SemGCN learns to capture semantic information such as local and global node relationships, which is not explicitly represented in the graph. These semantic relationships can be learned through end-to-end training from the ground truth without additional supervision or hand-crafted rules. We further investigate applying SemGCN to 3D human pose regression. Our formulation is intuitive and sufficient since both 2D and 3D human poses can be represented as a structured graph encoding the relationships between joints in the skeleton of a human body. We carry out comprehensive studies to validate our method. The results prove that SemGCN outperforms state of the art while using 90\% fewer parameters. The code can be found at \url{https://github.com/garyzhao/SemGCN}.
\end{abstract}

\section{Introduction}\label{sec:introduction}

Convolutional Neural Networks (CNNs) have successfully tackled classic computer vision problems such as image classification~\cite{elhoseiny2017link,krizhevsky2012imagenet,li2018joint,simonyan2015very}, object detection~\cite{he2017mask,ren2015faster,tang2018quantized,wang2015object,zhao2018pseudo,zhu2017multilayer} and generation~\cite{radford2016unsupervised,tian2018cr,han2017stackgan,zhao2018learning,zhu2018generative}, where the input image has a grid-like structure. However, many real-world tasks, \eg, molecular structures, social networks and 3D meshes, can only be represented in the form of irregular structures, where CNNs have limited applications.

In order to address this limitation, Graph Convolutional Networks (GCNs)~\cite{gori2005new,kipf2016semi,scarselli2009graph} have been introduced recently as a generalization of CNNs that can directly deal with a general class of graphs. They have achieved state-of-the-art performance when applied to 3D mesh deformation~\cite{ranjan2018generating,wang2018pixel}, image captioning~\cite{yao2018exploring}, scene understanding~\cite{yang2018graph}, and video recognition~\cite{wang2018videos,stgcn2018spatial}. These works utilize GCNs to model relations of visual objects for classification. In this paper, we investigate using deep GCNs for regression, which is another core problem of computer vision with many real-world applications.

However, GCNs cannot be directly applied to regression problems due to the following limitations in baseline methods~\cite{kipf2016semi,wang2018pixel,stgcn2018spatial}. First, to handle the issue that graph nodes may have various numbers of neighborhoods, the convolution filter shares the same weight matrix for all nodes, which is not comparable with CNNs. Second, previous methods are simplified by restricting the filters to operate in a one-step neighborhood around each node according to the guidance of~\cite{kipf2016semi}. The receptive field of the convolution kernel is limited to one due to this formulation, which severely impairs the efficiency of information exchanging especially when the network goes deeper.

In this work, we propose a novel graph neural network architecture for regression called \emph{Semantic Graph Convolutional Networks (SemGCN)} to address the above limitations. Specifically, we investigate learning semantic information encoded in a given graph, \ie, the local and global relations of nodes, which is not well-studied in previous works. SemGCN does not rely on hand-crafted constraints~\cite{du2015hierarchical,fang2018grammar,shahroudy2016ntu} to analyze the patterns for a specific application, and thus can be easily generalized to other tasks.

In particular, we study SemGCN for \emph{2D to 3D human pose regression}. Given a 2D human pose (and the optional relevant image) as input, we aim to predict the locations of its corresponding 3D joints in a certain coordinate space. Using SemGCN to formulate this problem is intuitive. Both 2D and 3D poses are able to be naturally represented by a canonical skeleton in the form of 2D or 3D coordinates, and SemGCN can explicitly exploit their spatial relations, which are crucial for understanding human actions~\cite{stgcn2018spatial}.

Our work makes the following contributions. First, we propose an improved graph convolution operation called \emph{Semantic Graph Convolution (SemGConv)} which is derived from CNNs. The key idea is to learn channel-wise weights for edges as priors implied in the graph, and then combine them with kernel matrices. This significantly improves the power of graph convolutions. Second, we introduce SemGCN where \textit{SemGConv} and non-local~\cite{wang2018non} layers are interleaved. This architecture captures both local and global relationships among nodes.  Third, we present an end-to-end learning framework to show that SemGCN can also incorporate external information, such as image content, to further boost the performance for 3D human pose regression.

The effectiveness of our approach is validated by comprehensive evaluation with a rigorous ablation study and comparisons with state of the art on standard 3D benchmarks. Our approach matches the performance of state-of-the-art techniques on Human3.6M~\cite{ionescu2014human3} using only 2D joint coordinates as inputs and 90\% fewer parameters. Meanwhile, our approach outperforms state of the art when incorporating image features. Furthermore, we also show the visual results of SemGCN, which demonstrate the effectiveness of our approach qualitatively. Note that the proposed framework can be easily generalized to other regression tasks, and we leave this for future work.

\section{Related Work}\label{sec:related_work}


\p{Graph convolutional networks.} Generalizing CNNs to inputs with graph-like structures is an important topic in the field of deep learning. In the literature, there have been several attempts to use recursive neural networks to process data represented in graph domains as directed acyclic graphs~\cite{frasconi1998general}. GNNs were introduced in~\cite{gori2005new,kipf2016semi,scarselli2009graph} as a more common solution to handle arbitrary graph data. The principle of constructing GCNs on graph generally follows two streams: the spectral perspective and the spatial perspective. Our work belongs to the second stream~\cite{kipf2016semi,niepert2016learning,velickovic2018graph}, where the convolution
filters are applied directly on the graph nodes and their neighbors.

Recent studies on computer vision have achieved state-of-the-art performance by leveraging GCNs to model the relations among visual objects~\cite{yang2018graph,yao2018exploring} or temporal sequences~\cite{wang2018videos,stgcn2018spatial}. This paper follows the spirit of them, while we explore applying GCNs for regression tasks, especially, 2D to 3D human pose regression.

\p{3D pose estimation.} Lee and Chen~\cite{lee1985determination} first investigated inferring 3D joints from their corresponding 2D projections. Later approaches either exploited nearest neighbors to refine the results of pose inference~\cite{gupta20143d,jiang20103d} or extracted hand-crafted features~\cite{agarwal2006recovering,ionescu2011latent,rogez2008randomized} for later regression. Other methods created over-complete bases which are suitable for representing human poses as sparse combinations~\cite{akhter2015pose,bogo2016keep,ramakrishna2012reconstructing,wang2014robust,zhou2016sparseness}. More and more studies focus on making use of deep neural networks to find the mapping between 2D and 3D joint locations. A couple of algorithms directly predicted 3D pose from the image~\cite{zhou2017towards}, while others combined 2D heatmaps with volumetric representation~\cite{pavlakos2017coarse}, pairwise distance matrix estimation~\cite{moreno20173d} or image cues~\cite{tekin2017learning} for 3D human pose regression.

Recently, it has been proven that 2D pose information is crucial for 3D pose estimation. Martinez~\etal~\cite{martinez2017simple} introduced a simple yet effective method which predicted 3D key points purely based on 2D detections. Fang~\etal~\cite{fang2018grammar} further extended this approach through pose grammar networks. These works focus on 2D to 3D pose regression, which are most relevant to the context of this paper.

Other methods use synthetic datasets which are generated from deforming a human template model with the ground truth~\cite{chen2016synthesizing,peng2018jointly,rogez2016mocap} or introduce loss functions involving high-level knowledge~\cite{park20183d,sun2017compositional,yang20183d} in addition to joints. They are complementary to the others. Remaining works target at exploiting temporal information~\cite{du2016marker,gupta20143d,hossain2018exploiting,tekin2016direct} for 3D pose regression. They are out of the scope of this paper, since we aim at handling the 2D pose from one single image. However, our method can be easily extended to sequence inputs, and we leave it for future work.

\section{Semantic Graph Convolutional Networks}\label{sec:sgcn}

We propose a novel graph network architecture to handle general regression tasks involving data that can be represented in the form of graphs. We first provide the background of GCNs and related baseline method. Then we introduce the detailed design of SemGCN.

We assume that graph data share the same topological structure, such as human skeletons~\cite{du2015hierarchical,ke2017new,vemulapalli2014human,stgcn2018spatial}, 3D morphable models~\cite{loper2015smpl,ranjan2018generating,zhao2019cartoonish} and citation networks~\cite{sen2008collective}. Other problems which own different graph structures in the same domain, \eg, protein-protein interaction~\cite{velickovic2018graph} and quantum chemistry~\cite{gilmer2017neural}, are out of the scope of this paper. This assumption makes it possible to learn priors implied in the graph structure, which motivates SemGCN.

\begin{figure*}
\begin{center}
\includegraphics[width=\linewidth]{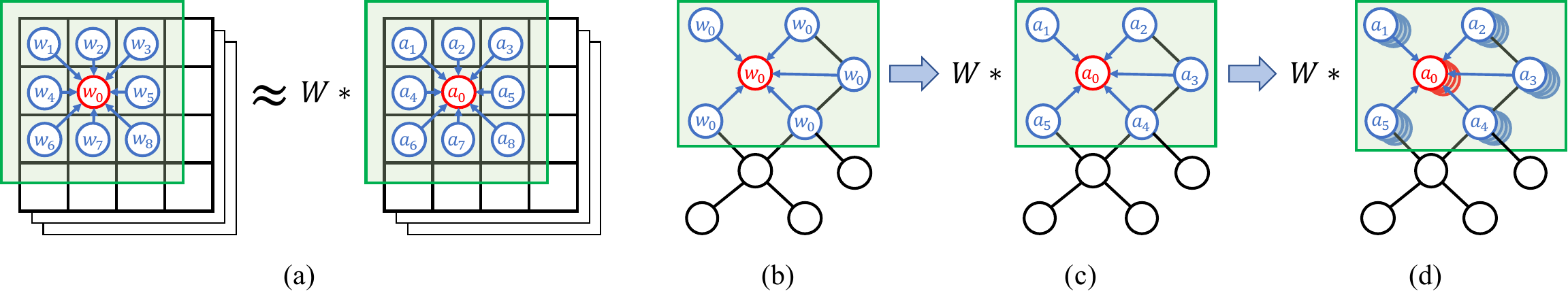}
\end{center}
\vspace{-12px}
   \caption{Illustration of the proposed Semantic Graph Convolutions. (a) The $3 \times 3$ convolution kernel of CNNs (highlighted in \textit{green}) learns a different transformation matrix $\boldsymbol{w}_i$ for each position inside the kernel. We approximate it by learning a weighting vector $\boldsymbol{a}_i$ for each position and a shared transformation matrix $\mathbf{W}$. (b) Conventional GCNs only learn a shared transformation matrix $\boldsymbol{w}_0$ for all nodes. (c) The approximated formulation in (a) can be directly extended to (b): we add an additional learnable weight $a_i$ for each node in the graph. (d) We further extend (c) to learn a channel-wise weighting vector $\boldsymbol{a}_i$ for each node. After combining them with the vanilla transformation matrix $\mathbf{W}$ in GCNs, we can obtain a new kernel operation for graphs which owns comparable learning capability with CNNs. The learned weight vectors show the local semantic relationships of neighboring nodes implied in the graph.}
\label{fig:graph_conv}
\end{figure*}

\subsection{ResGCN: A Baseline}\label{sec:baseline}

We will start by briefly recapping the `vanilla' GCNs as proposed in~\cite{kipf2016semi}. Let $\mathcal{G} = \{\mathbf{V},\mathbf{E}\}$ denote a graph where $\mathbf{V}$ is the set of $K$ nodes and $\mathbf{E}$ are edges, while $\overrightarrow{\boldsymbol{x}}^{(l)}_i \in \mathbb{R}^{D_l}$ and $\overrightarrow{\boldsymbol{x}}^{(l+1)}_i \in \mathbb{R}^{D_{l+1}}$ are the representations of node $i$ before and after the $l$-th convolution respectively. A graph based convolutional propagation can be applied to node $i$ in two steps. First, node representations are transformed by a learnable parameter matrix $\mathbf{W} \in \mathbb{R}^{D_{l+1} \times D_l}$. Second, these transformed node representations are gathered to node $i$ from its neighboring nodes $j \in \mathcal{N}(i)$, followed by a non-linear function (ReLU~\cite{nair2010rectified}). If node representations are collected into a matrix $\mathbf{X}^{(l)} \in \mathbb{R}^{D_l \times K}$, the convolution operation can be written as:
\begin{equation}
\label{eq:gc_m}
\mathbf{X}^{(l+1)} = \sigma\Big(\mathbf{W}\mathbf{X}^{(l)}\tilde{\mathbf{A}}\Big),
\end{equation}
where $\tilde{\mathbf{A}}$ is symmetrically normalized from $\mathbf{A}$ in conventional GCNs. $\mathbf{A} \in [0, 1]^{K \times K}$ is the adjacency matrix of $\mathcal{G}$, and we have $\alpha_{ij} = 1$ for node $j \in \mathcal{N}(i)$ and $\alpha_{ii} = 1$.

Wang~\etal~\cite{wang2018pixel} rephrased a very deep graph network based on Eq.~\ref{eq:gc_m} with residual connections~\cite{he2016deep} to learn the mapping between image features and 3D vertexes. We adopt its network architecture and treat it as our baseline which is denoted as ResGCN.

There are two clear drawbacks in Eq.~\ref{eq:gc_m}. First, in order to make the graph convolution work on nodes with arbitrary topologies, the learned kernel matrix $\mathbf{W}$ is shared for all edges. As a result, the relationships of neighboring nodes, or the internal structure in the graph, is not well exploited. Second, previous works only collect features from the first-order neighbors of each node. This is also limited because the receptive field is fixed to 1.

\subsection{Semantic Graph Convolutions}\label{sec:wgc}

\begin{figure*}
\begin{center}
\includegraphics[width=\linewidth]{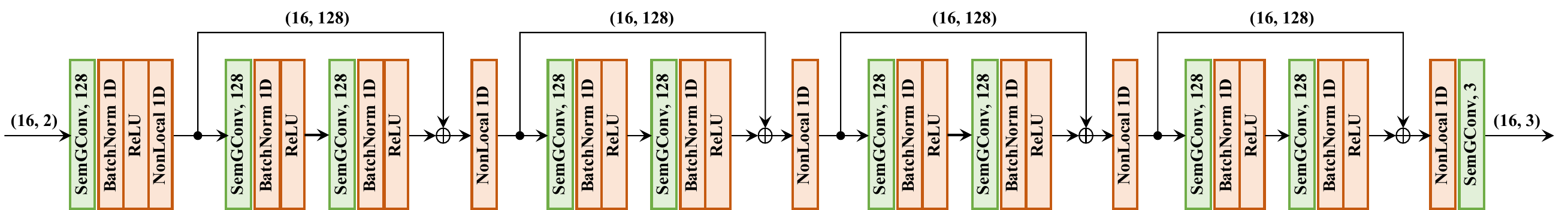}
\end{center}
\vspace{-0.1cm}
   \caption{Example of the proposed Semantic Graph Convolutional Networks. The building block of our network is one residual block~\cite{he2016deep} built by two \textit{SemGConv} layers with 128 channels, followed by one non-local layer~\cite{wang2018non}. This block is repeated four times. All \textit{SemGConv} layers are followed by batch normalization~\cite{ioffe2015batch} and a ReLU activation~\cite{nair2010rectified} except the last one.}
\label{fig:sgcn_architecture}
\end{figure*}

We show that learning semantic relationships of neighboring nodes implied in edges of the graph is effective to address the limitation of the shared kernel matrix.

The proposed approach builds on concepts from CNNs. Fig.~\ref{fig:graph_conv}(a) shows a CNN with a convolution kernel of size $3 \times 3$. It learns nine transformation matrices which are different from each other to encode features inside the kernel in the spatial dimension. This makes the operation own expressive power to model feature patterns contained in images. We find that this formulation can be approximated by learning a weighting vector $\overrightarrow{\boldsymbol{a}}_i$ for each position, and then combining them with a shared transformation matrix $\mathbf{W}$. If we represent the image feature map as a square grid graph whose nodes represent pixels, this approximated formulation can be directly extended to GCNs as shown in Fig.~\ref{fig:graph_conv}(c).

To this end, we propose \textit{Semantic Graph Convolution (SemGConv)}, where we add a learnable weighting matrix $\mathbf{M} \in \mathbb{R}^{K \times K}$ to conventional graph convolutions. And then Eq.~\ref{eq:gc_m} is transformed to:
\begin{equation}
\label{eq:ssgc_s}
\mathbf{X}^{(l+1)} = \sigma\Big(\mathbf{W}\mathbf{X}^{(l)}\rho_i\big(\mathbf{M}\odot\mathbf{A}\big)\Big),
\end{equation}
where $\rho_i$ is Softmax nonlinearity which normalizes the input matrix across all choices of node $i$; $\odot$ is an element-wise operation which returns $m_{ij}$ if $a_{ij} = 1$ or negatives with large exponents saturating to zero after $\rho_i$; $\mathbf{A}$ serves as a mask which forces that for node $i$ in the graph, we only compute the weights of its neighboring nodes $j \in \mathcal{N}(i)$.

As illustrated in Fig.~\ref{fig:graph_conv}(d), we can further extend Eq.~\ref{eq:ssgc_s} by learning a set of $\mathbf{M}_d \in \mathbb{R}^{K \times K}$, so that a different weighting matrix is applied to each channel $d$ of output node features:
\begin{equation}
\label{eq:ssgc_c}
\mathbf{X}^{(l+1)} = \mathop{\Big\|}^{D_{l+1}}_{d=1} \sigma\Big(\overrightarrow{\boldsymbol{w}}_d\mathbf{X}^{(l)}\rho_i\big(\mathbf{M}_d\odot\mathbf{A}\big)\Big),
\end{equation}
where $\|$ represents channel-wise concatenation, and $\overrightarrow{\boldsymbol{w}}_d$ is the $d$-th row of the transformation matrix $\mathbf{W}$.

\p{Comparison to previous GCNs.} Both aGCN~\cite{yang2018graph} and GAT~\cite{velickovic2018graph} follow a self-attention strategy~\cite{vaswani2017attention} to compute the hidden representations of each node in the graph by attending over its neighbors. They aim to estimate a weighting function \textit{depending on inputs} for edges to modulate information flow throughout the graph. By contrast, we target at learning \textit{input-independent} weights for edges which represent priors implied in the graph structures, \eg, how one joint influences other body parts in human pose estimation.

The edge importance weighting mask introduced in ST-GCN~\cite{stgcn2018spatial} is the most related work to ours but with following two sharp differences. First, no Softmax nonlinearity is leveraged after weighting by~\cite{stgcn2018spatial}, while we find it stabilizes the training and obtains better results, since the contributions of nodes to their neighbors are normalized by Softmax. Second, ST-GCN applies only one single learnable mask to all channels, but our Eq.~\ref{eq:ssgc_c} learns channel-wise different weights for edges. As a result, our model owns better capability to fit the data mapping.

\subsection{Network Architecture}\label{sec:architecture}

Capturing global and long-range relationships among nodes in the graph is able to efficiently address the problem of the limited receptive field. However, in order to maintain the behavior of GCNs, we restrict the feature updating mechanism by computing responses between nodes based on their representations other than learning new convolution filters. Therefore, we follow the non-local mean concept~\cite{buades2005non,wang2018non} and define the operation as:
\begin{equation}
\label{eq:sgcn_nl}
\overrightarrow{\boldsymbol{x}}^{(l+1)}_i = \overrightarrow{\boldsymbol{x}}^{(l)}_i + \frac{W_x}{K}\sum_{j = 1}^{K}f(\overrightarrow{\boldsymbol{x}}^{(l)}_i, \overrightarrow{\boldsymbol{x}}^{(l)}_j) \cdot g(\overrightarrow{\boldsymbol{x}}^{(l)}_j),
\end{equation}
where $W_x$ is initialized as zero; $f$ is a pairwise function to compute the affinity between node $i$ and all other $j$; $g$ computes the representation of the node $j$. In practice, Eq.~\ref{eq:sgcn_nl} can be implemented by the non-local layers proposed in~\cite{wang2018non}.

Based on Eq.~\ref{eq:ssgc_c} and~\ref{eq:sgcn_nl}, we propose a new network architecture for regression tasks called \textit{Semantic Graph Convolutional Networks}, where \textit{SemGConv} and non-local layers are interleaved to capture local and global semantic relations of nodes. Fig.~\ref{fig:sgcn_architecture} shows an example. In this work, SemGCN in all blocks has the same structure, which consists of one residual block~\cite{he2016deep} built by two \textit{SemGConv} layers with 128 channels, and then followed by one non-local layer. This block is repeated several times to make the network deeper. At the beginning of the network, one \textit{SemGConv} is used for mapping the inputs into the latent space; and we have an additional \textit{SemGConv} which projects the encoded features back to the output space. All \textit{SemGConv} layers are followed by batch normalization~\cite{ioffe2015batch} and a ReLU activation~\cite{nair2010rectified} except the last one. Note that if \textit{SemGConv} layers are replaced with vanilla graph convolutions and all non-local layers are removed, SemGCN downgrades to ResGCN in Sect.~\ref{sec:baseline}.

Intuitively, SemGCN can be regarded as a form of neural message passing system~\cite{gilmer2017neural} where the forward pass has two phases: messages are updated locally and then refined by the global state of the system. These two phases take turns to process messages so that the efficiency of information exchanging is improved for the whole system.

\section{3D Human Pose Regression}\label{sec:pose_regression}

\begin{figure*}
\begin{center}
\includegraphics[width=\linewidth]{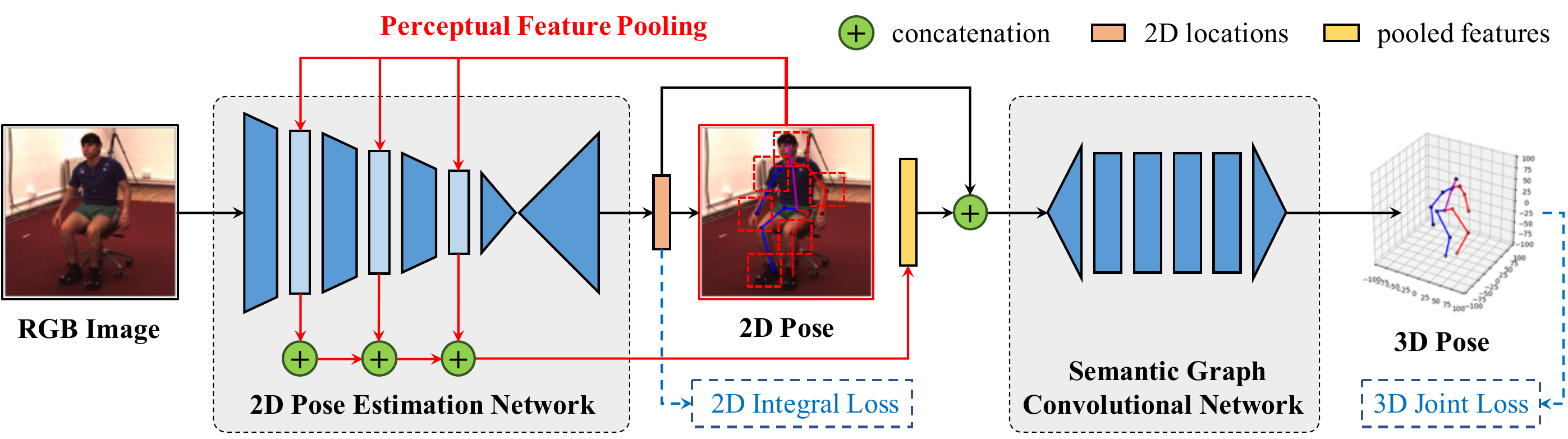}
\end{center}
\vspace{-0.2cm}
   \caption{Illustration of our framework incorporating image features for 3D human pose estimation. We pre-train a 2D pose estimation network to predict 2D joint locations. It also serves as a backbone network where we pool image features. The proposed SemGCN predicts 3D pose from 2D joints as well as image features. Note that the whole framework is end-to-end trainable.}
\label{fig:3dpose_architecture}
\end{figure*}

In this section, we present a novel end-to-end trainable framework which incorporates SemGCN in Sect.~\ref{sec:sgcn} with image features for 3D human pose regression.

\subsection{Framework Overview}\label{sec:overview}

Recently, it is proven that accurate 3D pose estimation can be directly obtained by using only 2D human poses as system inputs~\cite{martinez2017simple}. Formally, given a series of 2D joints $\mathbf{P} \in \mathbb{R}^{K \times 2}$ and their corresponding 3D joints $\mathbf{J} \in \mathbb{R}^{K \times 3}$ in a predefined camera coordinate system ($K$ is the numbers of joints), the system aims to learn a regression function $\mathcal{F}^*$ which minimizes the following error over a dataset containing $N$ human poses:
\begin{equation}
\label{eq:3dpose}
\mathcal{F}^* = \argmin_\mathcal{F}\frac{1}{N}\sum_{i = 1}^{N}\mathcal{L}(\mathcal{F}(\mathbf{P}_i), \mathbf{J}_i).
\end{equation}

We argue that image content is able to offer important cues for solving ambiguous cases, such as the classic turning ballerina optical illusion. Therefore, we extend Eq.~\ref{eq:3dpose} by treating image content as an additional constraint. The extended formulation can be denoted as:
\begin{equation}
\label{eq:3dpose_img}
\mathcal{F}^* = \argmin_\mathcal{F}\frac{1}{N}\sum_{i = 1}^{N}\mathcal{L}(\mathcal{F}(\mathbf{P}_i|I_i), \mathbf{J}_i),
\end{equation}
where $I_i$ is the image containing the aligned human pose of the 2D joints $\mathbf{P}_i$. In practice, $\mathbf{P}$ may be obtained as 2D ground truth locations under known camera parameters or from a 2D joint detector. In the latter case, the 2D detector has already encoded the perceptual features of the input image during the training process. This observation motivates the design of our framework.

An overview of our framework is shown in Fig.~\ref{fig:3dpose_architecture}. The whole framework consists of two neural networks. Given an image, one deep convolutional network is leveraged for 2D joints prediction; at the same time, it also serves as a backbone network and image features are pooled from its intermediate layers. Since 2D and 3D joint coordinates can be encoded in a human skeleton, the proposed SemGCN is used for automatically capturing the patterns embedded in the spatial configuration of the human joints. It predicts 3D coordinates according to the 2D pose as well as perceptual features from the backbone network.

Note that our framework effectively reduces to Eq.~\ref{eq:3dpose} when image features are not considered. As we demonstrate in experiments, SemGCN manages to effectively encode the mapping from 2D to 3D poses, and the performance can be further boosted when incorporating image content.

\subsection{Perceptual Feature Pooling}\label{sec:perceptual}

ResNet~\cite{he2016deep} and HourGlass~\cite{newell2016stacked} are widely adopted in conventional human pose detection problems. Empirically, we employ ResNet as the backbone network since its intermediate layers provide hierarchical features from images which are useful in computer vision problems such as object detection and segmentation~\cite{ren2015faster,zhao2018pseudo}.

Given the coordinate of each 2D joint in the input image, we pool features from multiple layers in ResNet. In particular, we concatenate features extracted from layer \textit{conv\_1} to \textit{conv\_4} using RoIAlign~\cite{he2017mask}. These perceptual features are then concatenated with the 2D coordinates and fed into SemGCN. Note that since all joints in the input image share the same scale, we pool the features in a squared bounding box centered on each joint with a fixed size, \ie, the mean bone length of the skeleton. This is illustrated in Fig.~\ref{fig:3dpose_architecture}.

\subsection{Loss Function}\label{sec:loss}

Most previous regression-based methods directly minimize the mean squared errors (MSE) of the predicted and ground truth joint positions~\cite{carreira2016human,martinez2017simple,tekin2016direct,zhou2016deep} or bone vectors~\cite{sun2017compositional}. Following the spirit of them, we employ a simple combination of joint and bone constraints in human poses as our loss function, which is defined as:
\begin{equation}
\label{eq:loss}
\mathcal{L}(\mathcal{B},\mathcal{J}) = \underbrace{\sum_{i = 1}^{M}||\tilde{\mathbf{B}}_i - \mathbf{B}_i||^2}_{\text{bone vectors}} + \underbrace{\sum_{i = 1}^{K}||\tilde{\mathbf{J}}_i - \mathbf{J}_i||^2}_{\text{joint positions}},
\end{equation}
where $\mathcal{J} = \{\tilde{\mathbf{J}}_i|i = 1, \dots, K\}$ are predicted 3D joint coordinates and $\mathcal{B} = \{\tilde{\mathbf{B}}_i|i = 1, \dots, M\}$ are bones computed from $\mathcal{J}$; $\mathbf{J}_i$ and $\mathbf{B}_i$ are corresponding ground truth in the dataset. Each bone is a directed vector pointing from the starting joint to its associated parent as defined in~\cite{sun2017compositional}.



\section{Experiments}\label{sec:experiments}

In this section, we first introduce settings and implementation details for evaluation, and then conduct an ablation study on components in our method, and finally report our results and comparisons with state-of-the-art methods.

\subsection{Implementation Details}\label{sec:implementation}

As suggested in the previous works~\cite{martinez2017simple,sun2017compositional,zhou2017towards}, it is impossible to train an algorithm to infer the 3D joint locations in an arbitrary coordinate space system. Therefore, we choose to predict 3D pose in the camera coordinate system~\cite{du2016marker,li2015maximum,pavlakos2017coarse,tekin2016direct}, which makes the 2D to 3D regression problem similar across different cameras.

We make use of the ground truth 2D joint locations provided in the dataset to align the 3D and 2D poses following the setting of~\cite{zhou2017towards}. This implies that we implicitly use the camera calibration information. Then, we zero-center both the 2D and 3D poses around the predefined root joint, \ie, the pelvis joint, which is in line with previous works and the standard protocol. Moreover, we do not use data augmentation during the training process for simplicity.

\p{Network training.} We use ResNet50 in~\cite{sun2017integral} as our backbone network, which is compatible with the integral loss and pre-trained on ImageNet~\cite{deng2009imagenet}. During training, we employ Adam~\cite{kingma2014adam} for optimization with a initial learning rate of 0.001 and use mini-batches of size 64. The learning rate is dropped with a decay rate of 0.5 when the loss on the validation set saturates. We initialize weights of the graph network using the initialization described in~\cite{glorot2010understanding}.

In our preliminary experiments, we observe that the direct end-to-end training of the whole network from scratch cannot achieve the best performance. We argue that this is likely because of the highly non-linear dependency between the graph network and conventional deep convolutional module for 2D pose estimation. Therefore, we utilize a multi-stage training scheme which is more stable and effective in practice. We first train the backbone network for 2D pose estimation from images using 2D ground truth. As described in~\cite{sun2017integral}, the integral loss is used. Then we fix the 2D pose estimation module and train the graph network for 2D to 3D pose regression using the output of 2D estimation module and the 3D ground truth. In this stage, the loss function defined in Eq.~\ref{eq:loss} is employed. At last, the whole network is fine-tuned with all data. Both integral loss and Eq.~\ref{eq:loss} are activated. Note that the final stage is end-to-end.

\subsection{Datasets and Evaluation Protocols}\label{sec:eval_datasets}

Our proposed approach is comprehensively evaluated on the most widely used dataset for 3D human pose estimation: Human3.6M~\cite{ionescu2014human3}, following the standard protocol.

\p{Datasets.} \textit{Human3.6M}~\cite{ionescu2014human3} is currently the largest publicly available dataset for 3D human pose estimation. This dataset contains 3.6 million of images captured by a MoCap System in an indoor environment, where 7 professional actors perform 15 everyday activities such as walking, eating, sitting, making a phone call and engaging in a discussion. Both 2D and 3D ground truth are available for supervised learning. Following the setting of~\cite{zhou2017towards}, the videos are down-sampled from 50fps to 10fps for both the training and testing sets to reduce redundancy. We also use \textit{MPII} dataset~\cite{andriluka20142d}, the state-of-the-art benchmark for 2D human pose estimation, for pre-training the 2D pose detector and qualitatively evaluation in the experiment.

\p{Evaluation protocols.} For Human3.6M~\cite{ionescu2014human3}, there are two common evaluation protocols using different training and testing data split in the literature. One standard protocol uses all 4 camera views in subjects S1, S5, S6, S7 and S8 for training and the same 4 camera views in subjects S9 and S11 for testing. Errors are calculated after the ground truth and predictions are aligned with the root joint. We refer to this as \textit{Protocol~\#1}. The other protocol makes use of six subjects S1, S5, S6, S7, S8 and S9 for training, and evaluation is performed on every 64th frame of S11. It also utilizes a rigid transformation to further align the predictions with the ground truth. This protocol is referred as \textit{Protocol~\#2}. In this work, we use Protocol \#1 in all the experiments for evaluation, since it is more challenging and matches the settings of our method.

The evaluation metric is the Mean Per Joint Position Error (MPJPE) in millimeter between the ground truth and the predicted 3D coordinates across all cameras and joints after aligning the pre-defined root joints (the pelvis joint). We use this metric for evaluation in the following sections.

Our network predicts the normalized locations of 3D joints. During testing, to calibrate the scale of the outputs, we require that the sum of length of all 3D bones is equal to that of a canonical skeleton as shown in~\cite{pavlakos2017coarse,zhou2017towards,zhou2018monocap}. Therefore, we follow the method in~\cite{zhou2017towards} for calibration.

\p{Configurations.} Our method is evaluated with the following two different configurations for 3D human pose estimation on Human3.6M.

\emph{Configuration \#1.} We only leverage 2D joints of the human pose as inputs. SemGCN in Sect.~\ref{sec:sgcn} is trained for regression and the \textit{SemGConv} layer defined in Eq.~\ref{eq:ssgc_s} is utilized. 2D ground truth (GT) or outputs from pre-trained 2D pose detectors are used for training and testing. In order to be in line with the setting of previous works~\cite{fang2018grammar,martinez2017simple}, we employ HourGlass~\cite{newell2016stacked} (HG) as the 2D detector. It is first pre-trained on MPII and then fine-tuned on Human3.6M. Only the joint loss in~Eq.~\ref{eq:loss} is employed.

\emph{Configuration \#2.} We use 2D images as inputs, and the proposed framework in Sect.~\ref{sec:pose_regression} is trained for regression. The channel-wise weighted \textit{SemGConv} defined in Eq.~\ref{eq:ssgc_c} is employed. ResNet50~\cite{he2016deep} is utilized as the backbone network for 2D pose estimation and feature pooling (RN w/ FP).

\subsection{Ablation Study}\label{sec:eval_ablation}

\begin{figure}
\begin{center}
\includegraphics[height=3.2cm]{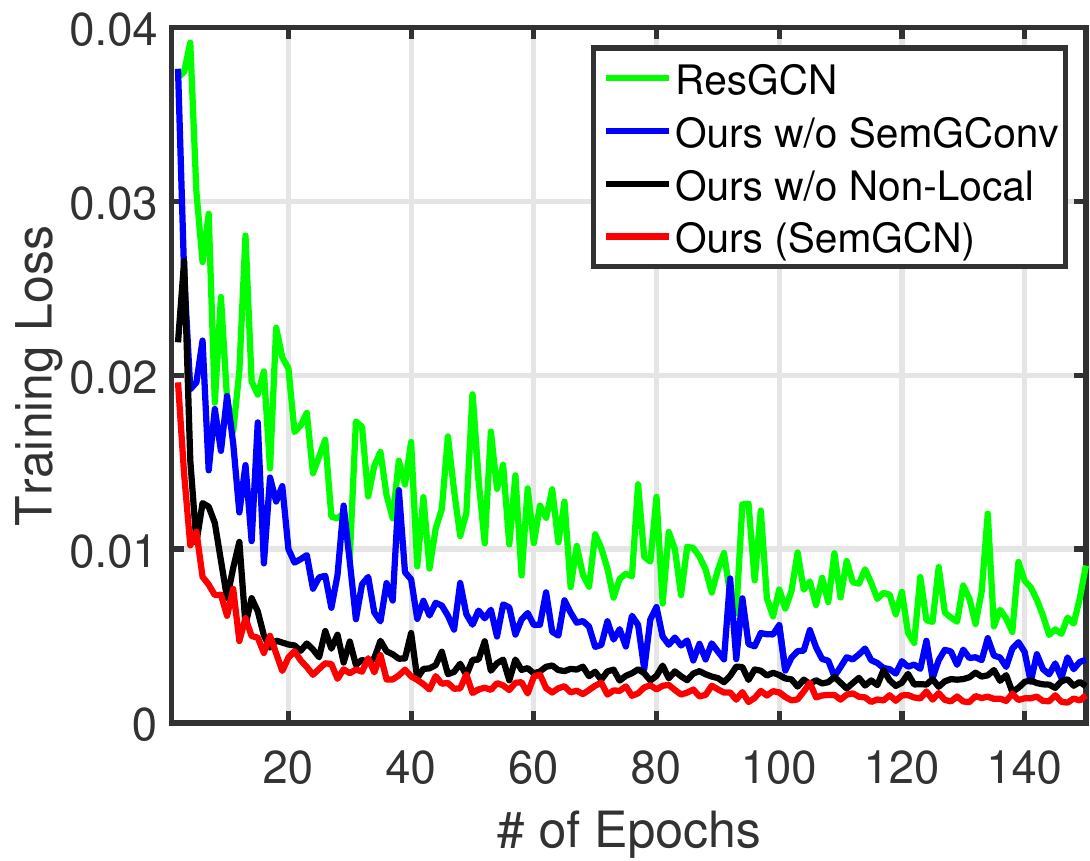}
\hspace{0.05cm}
\includegraphics[height=3.2cm]{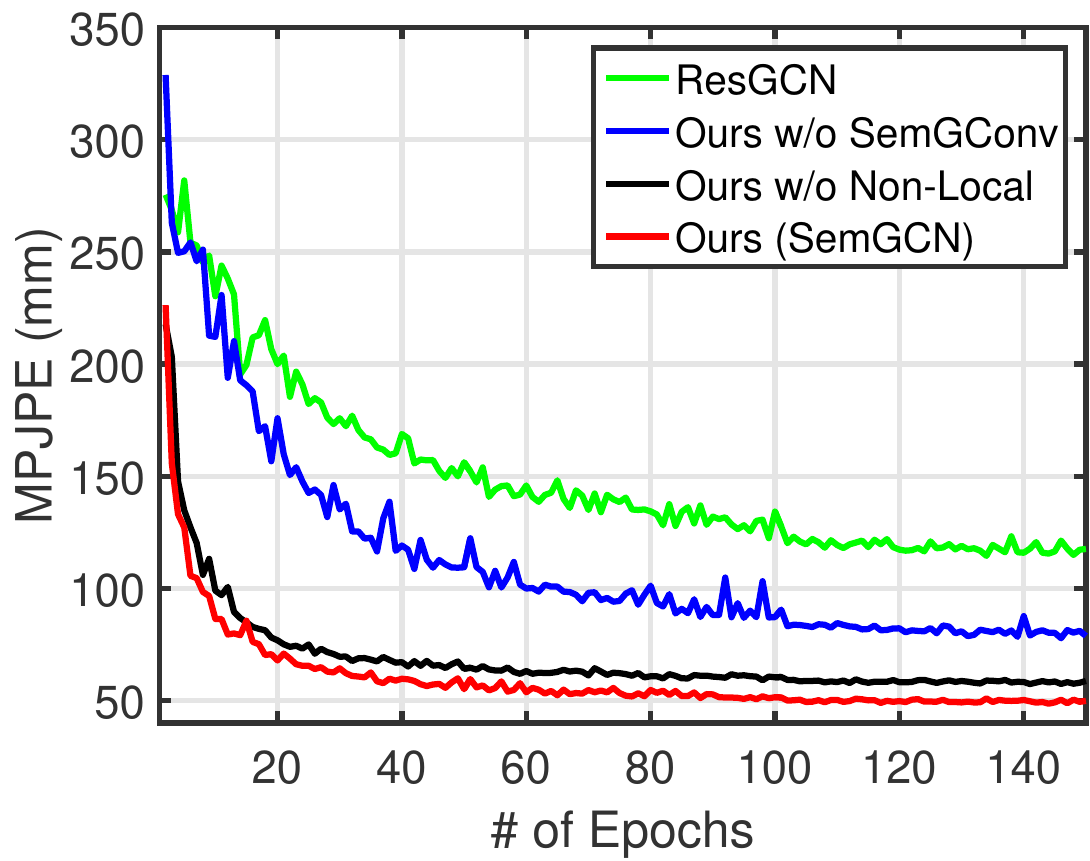}
\end{center}
\vspace{-0.15cm}
   \caption{Training curves (left) and testing errors (right) of our networks with different settings. Our full model has lower and smoother learning curves as well as better testing results.}
\label{fig:ablation}
\end{figure}

\begin{table}
\begin{center}
\begin{tabular}{l|c|c}
\hline
Method & \# of params & MPJPE (mm)\\
\hline\hline
ResGCN & 0.14M & 94.4\\
Ours w/o SemGConv & 0.30M & 65.9\\
Ours w/o Non-Local & 0.27M & 52.5\\
Ours (SemGCN) & 0.43M & 43.8\\
\hline
\end{tabular}
\end{center}
\caption{2D to 3D pose regression errors and the parameter numbers of our networks with different settings on Human3.6M dataset~\cite{ionescu2014human3}. Our full model achieves the best performance.}
\label{tbl:ablation}
\end{table}

\begin{table*}
\begin{center}
\resizebox{\textwidth}{!}{%
\begin{tabular}{@{}lrrrrrrrrrrrrrrrr@{}}
\hline\noalign{\smallskip}
\textbf{Protocol \#1} & Direct. & Discuss & Eating & Greet & Phone & Photo & Pose & Purch. & Sitting & SittingD. & Smoke & Wait & WalkD. & Walk & WalkT. & Avg. \\
\noalign{\smallskip}\hline\hline\noalign{\smallskip}
Ionescu~\etal~\cite{ionescu2014human3} PAMI'16 & 132.7 & 183.6 & 132.3 & 164.4 & 162.1 & 205.9 & 150.6 & 171.3 & 151.6 & 243.0 & 162.1 & 170.7 & 177.1 & 96.6 & 127.9 & 162.1\\
Tekin~\etal~\cite{tekin2016direct} CVPR'16 & 102.4 & 147.2 & 88.8 & 125.3 & 118.0 & 182.7 & 112.4 & 129.2 & 138.9 & 224.9 & 118.4 & 138.8 & 126.3 & 55.1 & 65.8 & 125.0\\
Zhou~\etal~\cite{zhou2016sparseness} CVPR'16 & 87.4 & 109.3 & 87.1 & 103.2 & 116.2 & 143.3 & 106.9 & 99.8 & 124.5 & 199.2 & 107.4 & 118.1 & 114.2 & 79.4 & 97.7 & 113.0\\
Du~\etal~\cite{du2016marker} ECCV'16 & 85.1 & 112.7 & 104.9 & 122.1 & 139.1 & 135.9 & 105.9 & 166.2 & 117.5 & 226.9 & 120.0 & 117.7 & 137.4 & 99.3 & 106.5 & 126.5\\
Chen \& Ramanan~\cite{chen20173d} CVPR'17 & 89.9 & 97.6 & 89.9 & 107.9 & 107.3 & 139.2 & 93.6 & 136.0 & 133.1 & 240.1 & 106.6 & 106.2 & 87.0 & 114.0 & 90.5 & 114.1\\
Pavlakos~\etal~\cite{pavlakos2017coarse} CVPR'17 & 67.4 & 71.9 & 66.7 & 69.1 & 72.0 & 77.0 & 65.0 & 68.3 & 83.7 & 96.5 & 71.7 & 65.8 & 74.9 & 59.1 & 63.2 & 71.9\\
Mehta~\etal~\cite{mehta2017monocular} 3DV'17 & 52.6 & 64.1 & 55.2 & 62.2 & 71.6 & 79.5 & 52.8 & 68.6 & 91.8 & 118.4 & 65.7 & 63.5 & 49.4 & 76.4 & 53.5 & 68.6\\
Zhou~\etal~\cite{zhou2017towards} ICCV'17 & 54.8 & 60.7 & 58.2 & 71.4 & 62.0 & 65.5 & 53.8 & 55.6 & 75.2 & 111.6 & 64.1 & 66.0 & 51.4 & 63.2 & 55.3 & 64.9\\
Martinez~\etal~\cite{martinez2017simple} ICCV'17 & 51.8 & 56.2 & 58.1 & 59.0 & 69.5 & 78.4 & 55.2 & 58.1 & 74.0 & 94.6 & 62.3 & 59.1 & 65.1 & 49.5 & 52.4 & 62.9\\

Sun~\etal~\cite{sun2017compositional} ICCV'17 & 52.8 & 54.8 & 54.2 & \bf{54.3} & \underline{61.8} & \underline{53.1} & 53.6 & 71.7 & 86.7 & 61.5 & 67.2 & \bf{53.4} & 47.1 & 61.6 & 53.4 & 59.1\\

Fang~\etal~\cite{fang2018grammar} AAAI'18 & 50.1 & \underline{54.3} & 57.0 & 57.1 & 66.6 & 73.3 & 53.4 & 55.7 & 72.8 & 88.6 & 60.3 & 57.7 & 62.7 & \underline{47.5} & 50.6 & 60.4\\
Yang~\etal~\cite{yang20183d} CVPR'18 & 51.5 & 58.9 & \bf{50.4} & 57.0 & 62.1 & 65.4 & \underline{49.8} & \underline{52.7} & \underline{69.2} & 85.2 & \bf{57.4} & 58.4 & \underline{43.6} & 60.1 & \underline{47.7} & 58.6\\


Hossain \& Little~\cite{hossain2018exploiting} ECCV'18 & 48.4 & \bf{50.7} & 57.2 & \underline{55.2} & 63.1 & 72.6 & 53.0 & \bf{51.7} & \bf{66.1} & 80.9 & \underline{59.0} & \underline{57.3} & 62.4 & \bf{46.6} & 49.6 & \underline{58.3}\\

\noalign{\smallskip}\hline\noalign{\smallskip}
Ours (HG) & \underline{48.2} & 60.8 & 51.8 & 64.0 & 64.6 & 53.6 & 51.1 & 67.4 & 88.7 & \underline{57.7} & 73.2 & 65.6 & 48.9 & 64.8 & 51.9 & 60.8\\
Ours (RN w/ FP) & \bf{47.3} & 60.7 & \underline{51.4} & 60.5 & \bf{61.1} & \bf{49.9} & \bf{47.3} & 68.1 & 86.2 & \bf{55.0} & 67.8 & 61.0 & \bf{42.1} & 60.6 & \bf{45.3} & \bf{57.6}\\
\noalign{\smallskip}\hline\noalign{\smallskip}
Ours (GT) & 37.8 & 49.4 & 37.6 & 40.9 & 45.1 & 41.4 & 40.1 & 48.3 & 50.1 & 42.2 & 53.5 & 44.3 & 40.5 & 47.3 & 39.0 & 43.8\\
\noalign{\smallskip}\hline
\end{tabular}}
\end{center}
\caption{Quantitative comparisons of  Mean Per Joint Position Error (mm) between the estimated pose and the ground truth on Human3.6M~\cite{ionescu2014human3} under Protocol \#1. We show the results of our model (Sect.~\ref{sec:sgcn}) trained and tested with the 2D predictions of HourGlass~\cite{newell2016stacked} (HG) as inputs using Configuration \#1, and the results of our network presented in Sect.~\ref{sec:pose_regression} which incorporate image features (RN w/ FP) during training and testing under Configuration \#2. We also show an upper bound of our method which uses 2D ground truth (GT) as the input for training and testing. The top two best methods of each action are highlighted in bold and underlined respectively.
}
\label{tbl:h36m}
\end{table*}

\begin{table}
\begin{center}
\begin{tabular}{l|c|c}
\hline
Method & \# of params & MPJPE (mm)\\
\hline\hline
aGCN~\cite{yang2018graph} / GAT~\cite{velickovic2018graph} & 0.16M & 82.9\\
ST-GCN~\cite{stgcn2018spatial} & 0.27M & 57.4\\
FC~\cite{martinez2017simple} & 4.29M & 45.5 (62.9)\\
FC~\cite{martinez2017simple} w/ PG~\cite{fang2018grammar} & - & 43.3 (60.4)\\
\hline
Ours & 0.43M & 43.8 (60.8)\\
Ours w/ PG~\cite{fang2018grammar} & - & 42.5 (59.8)\\
\hline
\end{tabular}
\end{center}
\caption{Evaluation of 2D to 3D pose regression on Human3.6M datasets~\cite{ionescu2014human3}. Errors within the parentheses are computed by using the 2D estimations from HG~\cite{newell2016stacked} as inputs during training and testing. Otherwise, 2D ground truth is utilized. Our method advances other GCN-based approaches by 20\% and achieves the state-of-the-art performance using 90\% fewer parameters than~\cite{martinez2017simple}.}
\label{tbl:2dto3d}
\end{table}

We conduct the ablation study on the proposed method in Sect.~\ref{sec:sgcn}. Configuration \#1 is employed. Our SemGCN consists of two main components: \textit{SemGConv} and non-local layers. To verify them, we train two variants of SemGCN: one only uses \textit{SemGConv} and the other only uses non-local layers. Then we evaluate them together with the baseline method in Sect.~\ref{sec:baseline} (ResGCN) and our full model in Sect.~\ref{sec:architecture} on Human3.6M. Note that in order to get rid of the influence from the 2D pose detector, we report the results using 2D ground truth for training and testing.

All models are trained based on the architecture shown in Fig.~\ref{fig:sgcn_architecture} after 200 epochs. Results are shown in Table~\ref{tbl:ablation}. We also show their curves of training losses and testing errors in Fig.~\ref{fig:ablation}. We can see that our model with more components performs better than those with fewer components, which indicates the efficacy of each part of our algorithm. Moreover, our networks with \textit{SemGConv} have much smoother training curves which demonstrates that learning local relations among nodes stabilizes the training process as well.

\begin{figure*}
\begin{center}
\includegraphics[width=\linewidth]{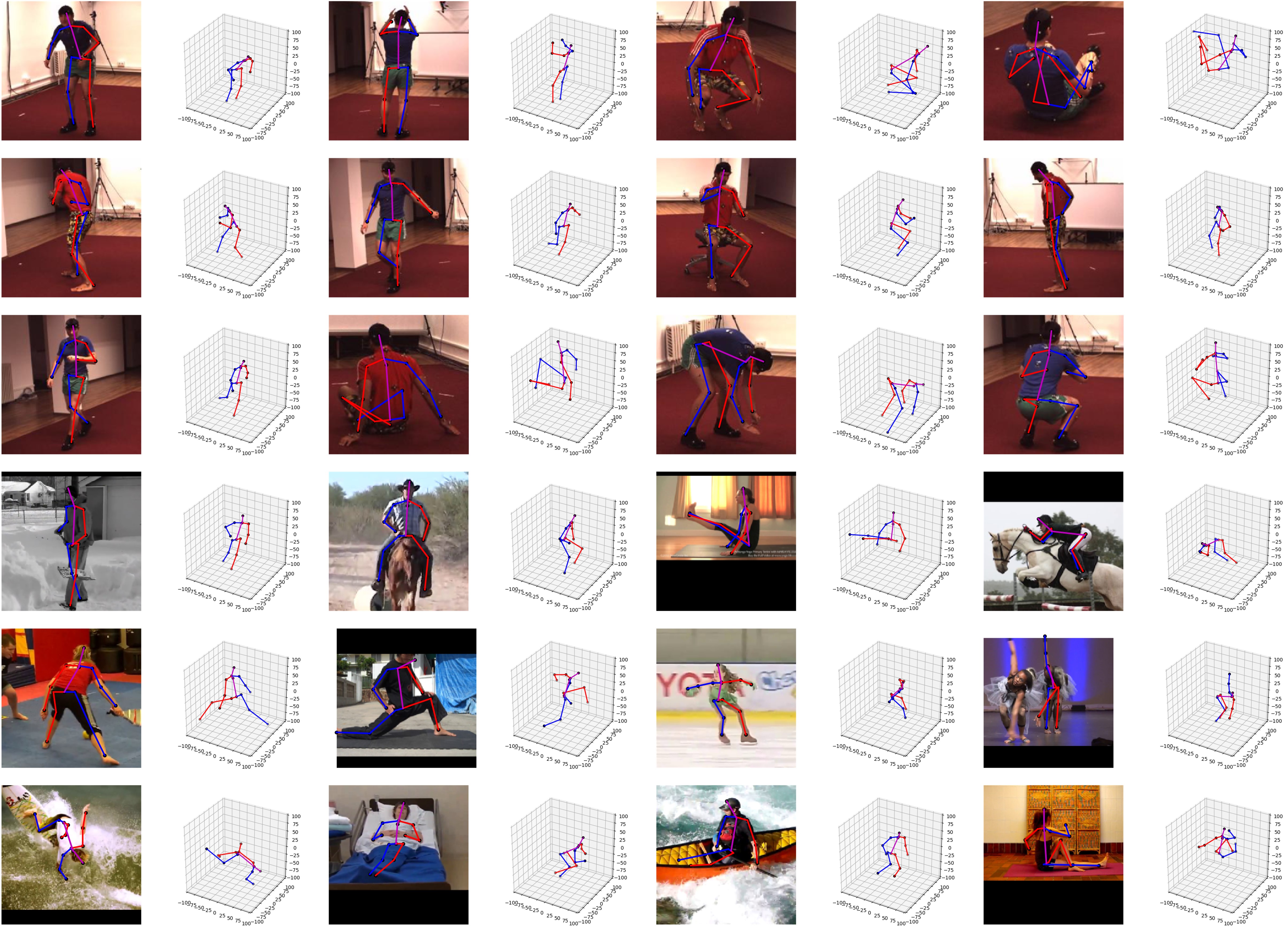}
\end{center}
   \caption{Visual results of our method on Human3.6M~\cite{ionescu2014human3} and MPII~\cite{andriluka20142d}. The first three rows show results on Human3.6M. Results on MPII are drawn in the last three rows. The bottom row shows four typical failure cases. Best viewed in color.}
\label{fig:qualitative}
\end{figure*}

\subsection{Evaluation on 3D Human Pose Regression}\label{sec:eval_3dpose}

\p{2D to 3D pose regression.} We first evaluate our method for 2D to 3D pose regression and only Configuration \#1 is leveraged. We compared ours with three GCN-based methods: aGCN~\cite{yang2018graph}, GAT~\cite{velickovic2018graph} and ST-GCN~\cite{stgcn2018spatial}, and two state-of-the-art approaches: FC~\cite{martinez2017simple} and PG~\cite{fang2018grammar}. As ST-GCN~\cite{stgcn2018spatial} is designed for videos, we set its temporal dimension to one for images. PG proposed a framework to refine the 3D pose, which is complementary to FC and ours. Therefore, we also report our results refined by PG.

The results are reported in Table~\ref{tbl:2dto3d}. Our approach outperforms other GCN-based approaches by a large margin (about 20\%). More importantly, our method achieves the state-of-the-art performance with around 90\% fewer parameters than~\cite{martinez2017simple}. Meanwhile, the runtime of SemGCN reduces 10\% compared with~\cite{martinez2017simple}, which is around 1.8ms for a forward pass on a Titan Xp GPU. After we refined our results by PG, our approach obtains the best performance.

\p{Comparison with the state of the art.} We show evaluation results under Configuration \#1 and \#2. Note that many leading methods have sophisticated frameworks or learning strategies. Some of them aim at in-the-wild images~\cite{sun2017integral,yang20183d,zhou2017towards} or exploit temporal information~\cite{du2016marker,gupta20143d,hossain2018exploiting,tekin2016direct}, while some other approaches use complex loss functions~\cite{sun2017compositional,yang20183d}. These methods are with different research targets compared to ours. Therefore, we include some of them during evaluation for completeness. Table~\ref{tbl:h36m} reports the results.

We find that our method using only 2D joints as inputs is able to match the state-of-the-art performance. After incorporating image features, our network sets the new state of the art. Especially, we improve previous methods by a large margin for the action of directions, taking photo, posing, sitting down, walking dog and walking together. We hypothesize that this is due to the severe self-occlusions in these actions, while they can be effectively encoded by our SemGCN using relations within graphs. The result of our method trained and tested with ground truth 2D joint locations shows our upper bound.

\p{Qualitative results.} In Fig.~\ref{fig:qualitative}, we show the visual results of our method on Human3.6M and the test set of MPII. MPII contains in-the-wild images with novel human poses which are not similar to the examples in Human3.6M. As seen, our method is able to accurately predict 3D pose for both indoor and most in-the-wild images. It indicates that SemGCN can effectively encode relationships among joints and further generalize them to some novel cases.

The bottom row of Fig.~\ref{fig:qualitative} also shows typical failure cases of our method. These images include extreme poses which are largely different from those in Human3.6M. Our method failed to handle them but still yields reasonable 3D poses.

\section{Conclusions}\label{sec:conclusions}

We present a novel model for 3D human pose regression, the Semantic Graph Convolutional Networks (SemGCN). Our method has addressed the key challenges of GCNs by learning local and global semantic relations among nodes in the graph. The combination of SemGCN and features pooled from image content further improves the performance in 3D human pose estimation. Comprehensive evaluation results show that our network obtains state-of-the-art performance with 90\% fewer parameters compared with the closest work. The proposed SemGCN also opens up many possible directions for future works. For example, how to incorporate temporal information, such as videos, into SemGCN becomes a natural question.

\vspace{4px}\noindent\textbf{Acknowledgments.}
This work was funded partly by grant BAAAFOSR-2013-0001 to Dimitris Metaxas. This work was also partly supported by NSF 1763523, 1747778, 1733843 and 1703883 Awards. Mubbasir Kapadia was funded partly by NSF IIS-1703883, NSF S\&AS-1723869, and DARPA SocialSim-W911NF-17-C-0098.

{\small
\bibliographystyle{ieee_fullname}
\bibliography{egbib}
}

\clearpage
\begin{appendices}

\section{Supplementary Material}\label{sec:supp}

This supplementary material provides additional results supporting the claims of the main paper. First, we provide more details about Semantic Graph Convolutional Networks (SemGCN), including the skeleton representation for building the graph (Sect.~\ref{sec:skeleton}) and the implementation of graph convolutions (Sect.~\ref{sec:gconv}) and non-local layers (Sect.~\ref{sec:layers}). Additionally, to better understand the proposed Semantic Graph Convolutions, we provide the visualization results of the learned weights implied in the graph after training (Sect.~\ref{sec:sgcs}).

\subsection{Skeleton Representation}\label{sec:skeleton}

Following the setting of previous works~\cite{fang2018grammar,martinez2017simple,sun2017compositional,sun2017integral,zhou2017towards}, we utilize a common human skeleton representation for Human3.6M~\cite{ionescu2014human3} and MPII~\cite{andriluka20142d} to build the graph of SemGCN. This skeleton is visualized in Fig.~\ref{fig:skeleton}(left). It consists of 16 joints and we define the \emph{pelvis} joint as the root joint. Note that the skeleton is initialized as an \emph{undirected} graph in SemGCN before training. After we finish training the network, it will transform to a \emph{weighted directed} graph represented by $\rho_i\big(\mathbf{M}\odot\mathbf{A}\big)$ in Eq.~\ref{eq:ssgc_s} and \ref{eq:ssgc_c}.

In Fig.~\ref{fig:skeleton}(left), we also show the bone vectors we employed in Eq.~\ref{eq:loss} to compute the bone loss. Let the bone $\mathbf{B}_k$ be directed from the joint $\mathbf{J}_{parent(k)}$ to the target joint $\mathbf{J}_k$, and we define the bone vector as:
\begin{equation}
\label{eq:bone}
\mathbf{B}_k = \mathbf{J}_{parent(k)} - \mathbf{J}_k.
\end{equation}
This formulation is consistency with~\cite{sun2017compositional}. However, in order to be in line with the setting of previous works~\cite{fang2018grammar,martinez2017simple} for fair comparison, the bone loss is not employed in \emph{Configuration \#1} of the experiments.

\subsection{Implementation of Graph Convolutions}\label{sec:gconv}

Some previous approaches~\cite{wang2018pixel,stgcn2018spatial} proposed to leverage two different transformation matrixes other than one in the graph convolutions. To be specific, when the graph convolutional filter is applied to node $i$ in the graph, one matrix $\mathbf{W}_0$ is employed to transform the  representation of node $i$ while the other matrix $\mathbf{W}_1$ is learned for all its neighbors. According to this formulation, we rewrite Eq.~\ref{eq:gc_m} to:
\begin{equation}
\label{eq:gc_m0}
\mathbf{X}^{(l+1)} = \sigma\Big(\mathbf{I} \otimes \mathbf{W}_0\mathbf{X}^{(l)}\tilde{\mathbf{A}} + \big(\mathbf{1} - \mathbf{I}\big) \otimes \mathbf{W}_1\mathbf{X}^{(l)}\tilde{\mathbf{A}}\Big),
\end{equation}
where $\otimes$ denotes element-wise multiplication and $\mathbf{I}$ is the identity matrix. We also implement the proposed \textit{SemGConv} defined by Eq.~\ref{eq:ssgc_s} and \ref{eq:ssgc_c} in the similar manner.

\subsection{Non-local Layers}\label{sec:layers}

We follow the guidance of Wang~\etal~\cite{wang2018non} to implement the non-local layers in SemGCN. For computational efficiency, we down-sample both the feature dimension and number of nodes in the graph when calculating the embedding of $f(\overrightarrow{\boldsymbol{x}}^{(l)}_i, \overrightarrow{\boldsymbol{x}}^{(l)}_j)$ in Eq.~\ref{eq:sgcn_nl}.

\p{Feature embedding.} We use ``concatenation'' for the implementation of $f$. Two mapping functions $\theta(\cdot)$ and $\phi(\cdot)$ are employed to down-sample the feature of each node from 128 to 64 channels. They are implemented as convolutions with the kernel size of 1. Then we define $f$ as:
\begin{equation}
\label{eq:embedding}
f(\overrightarrow{\boldsymbol{x}}^{(l)}_i, \overrightarrow{\boldsymbol{x}}^{(l)}_j) = \text{ReLU}(\mathbf{w}_f[\theta(\overrightarrow{\boldsymbol{x}}^{(l)}_i) \| \phi(\overrightarrow{\boldsymbol{x}}^{(l)}_j)]),
\end{equation}
where $[\cdot \| \cdot]$ denotes concatenation, and $\mathbf{w}_f$ is the parameters to be learned to project the concatenated vector to a scalar.

\begin{figure}
\begin{center}
\includegraphics[width=\linewidth]{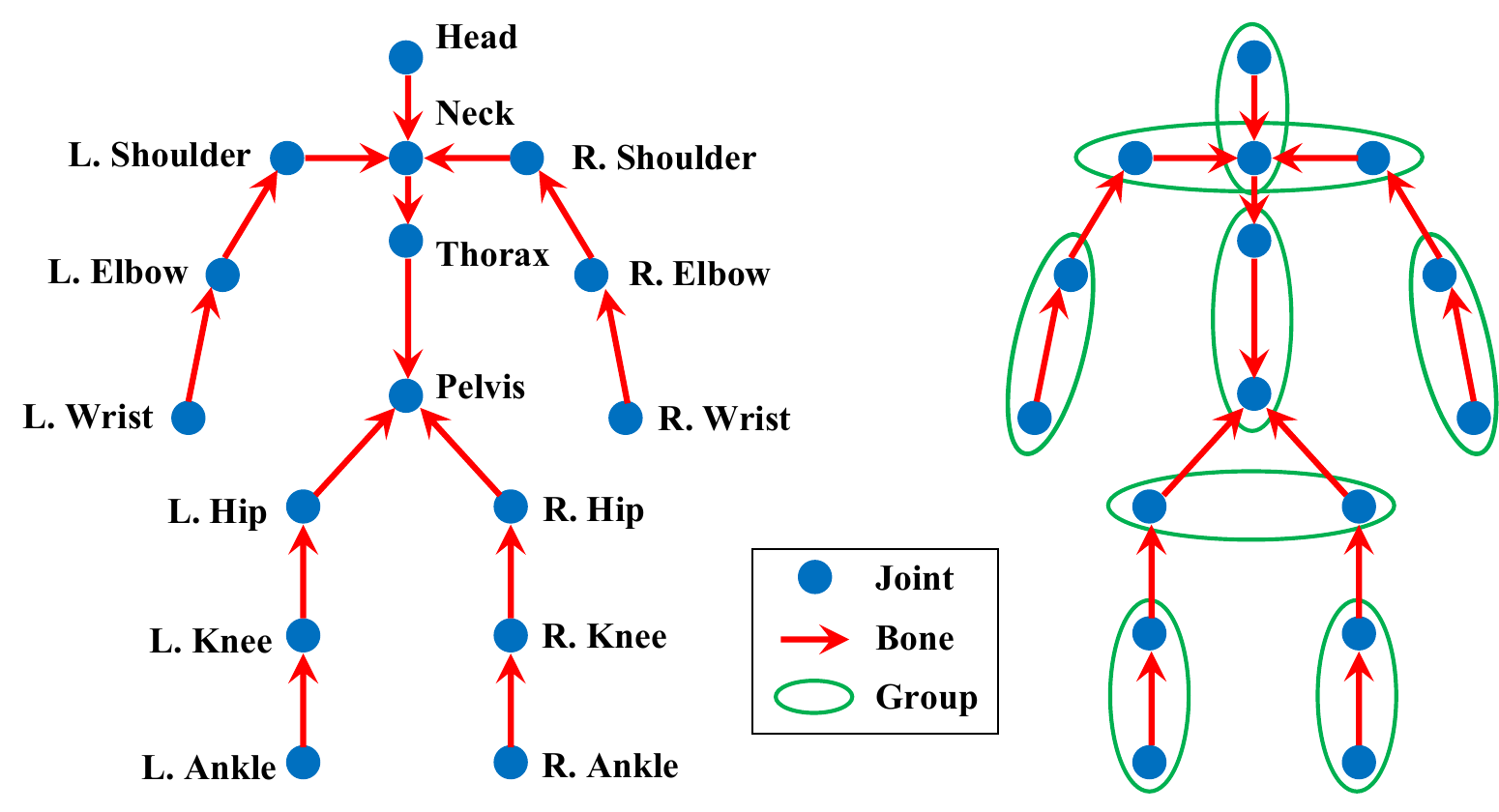}
\end{center}
   \caption{Left: the template human skeleton we utilized to build the graph. Each red arrow represents the bone vector for computing the loss function. Right: the grouping (down-sampling) strategy of the human skeleton we employed in the non-local layers.}
\label{fig:skeleton}
\end{figure}

\p{Node grouping.} We also use max pooling to down-sample the number of nodes in the graph. Fig~\ref{fig:skeleton}(right) illustrates the grouping strategy we employed. The number of nodes contained in the graph reduces from 16 to 8 after the max pooling operation. This strategy is used for all non-local layers in SemGCN. In the experiments, we find that this pooling operation can speed up the runtime, while it does not influence the final accuracy of the regression.

\subsection{Visualization of Weights in SemGCN}\label{sec:sgcs}

\begin{figure*}
\begin{center}
\includegraphics[height=3.5cm,align=t]{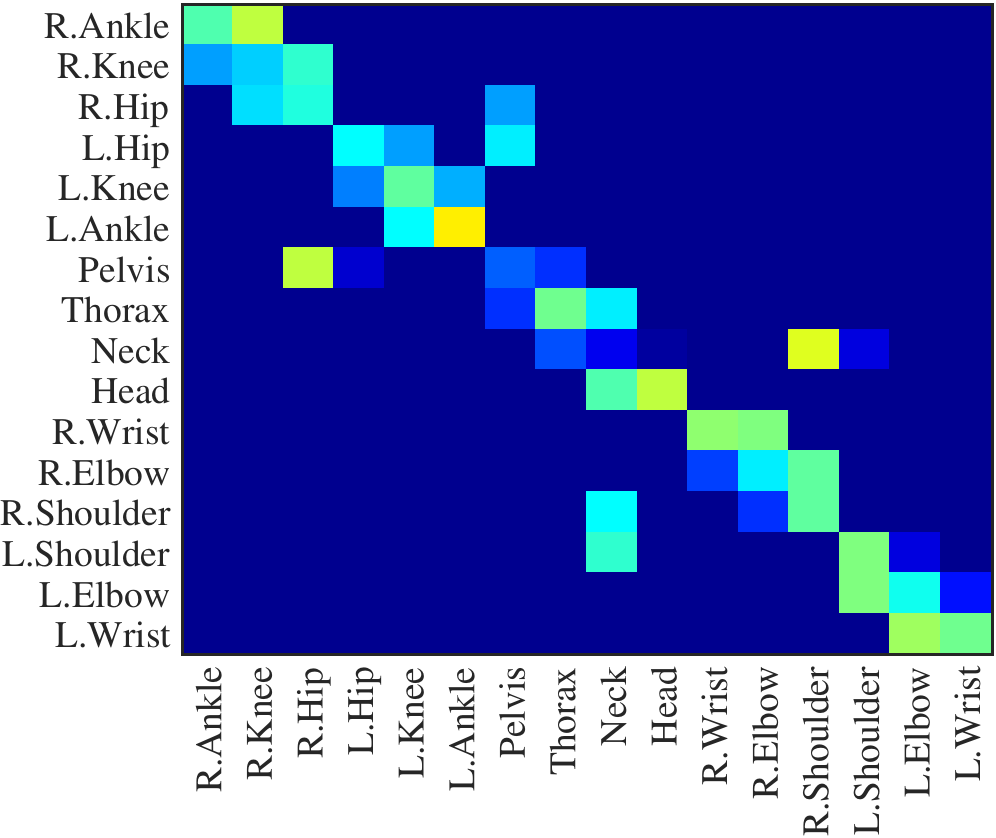}
\includegraphics[height=3.5cm,align=t]{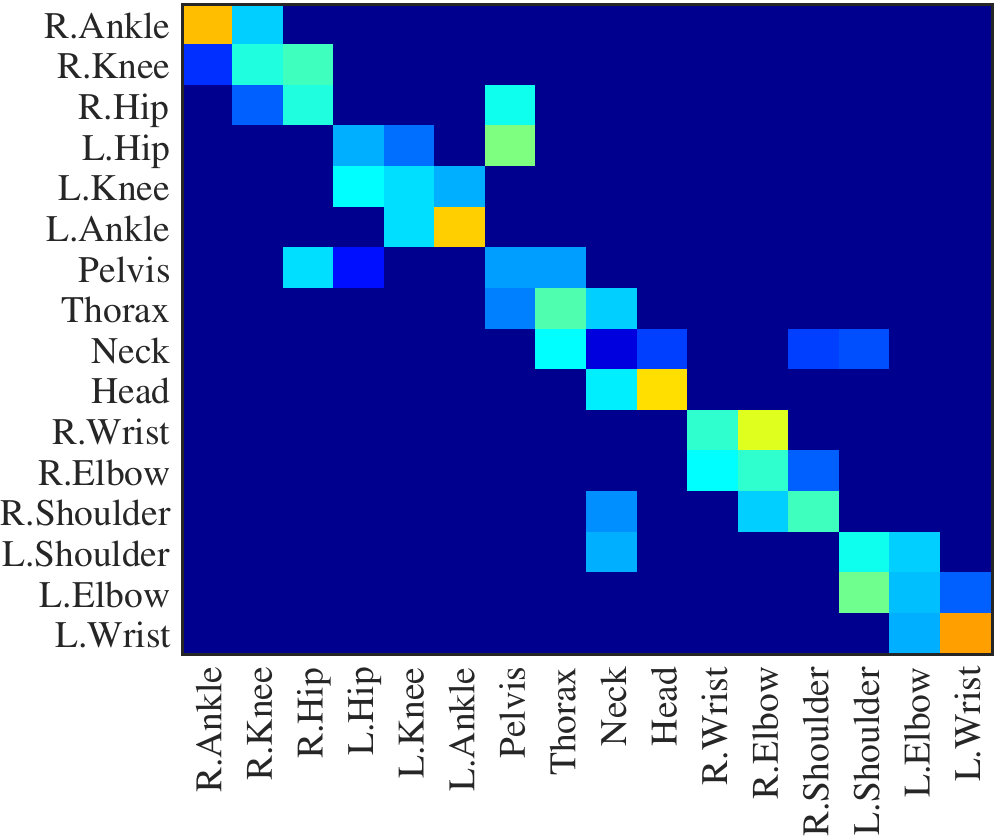}
\includegraphics[height=3.5cm,align=t]{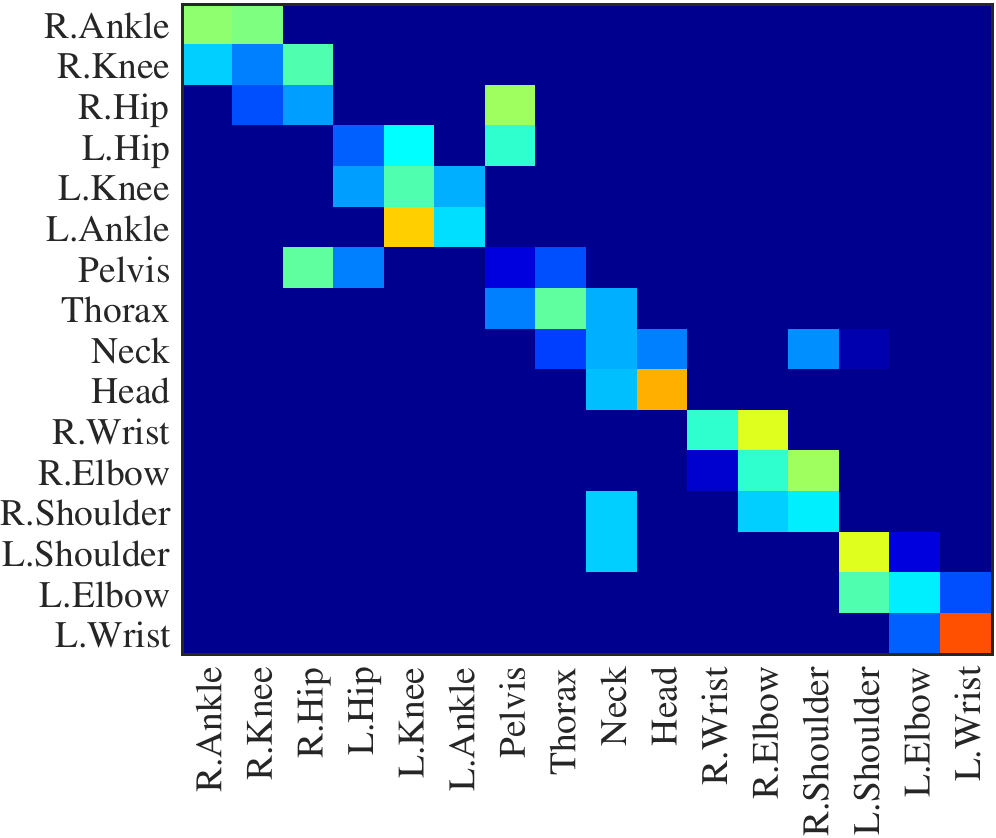}
\includegraphics[height=3.5cm,align=t]{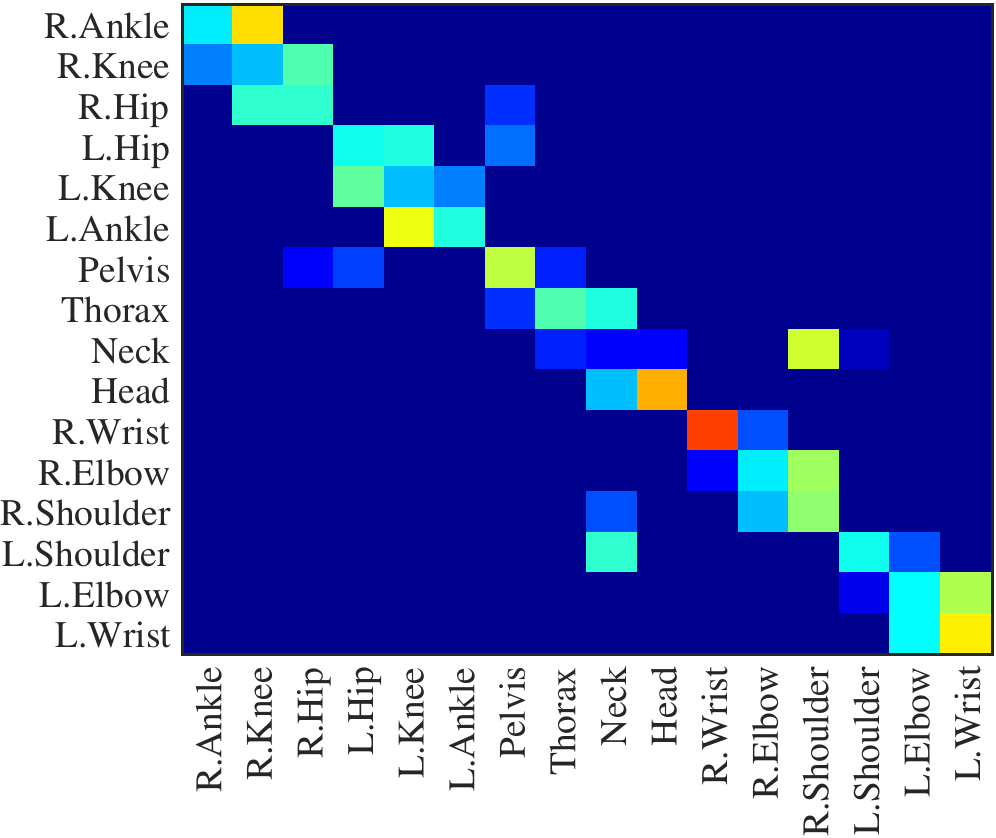}
\includegraphics[height=2.4cm,align=t]{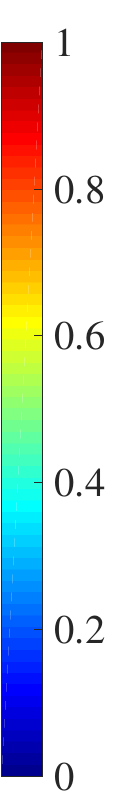}
\includegraphics[height=3.5cm,align=t]{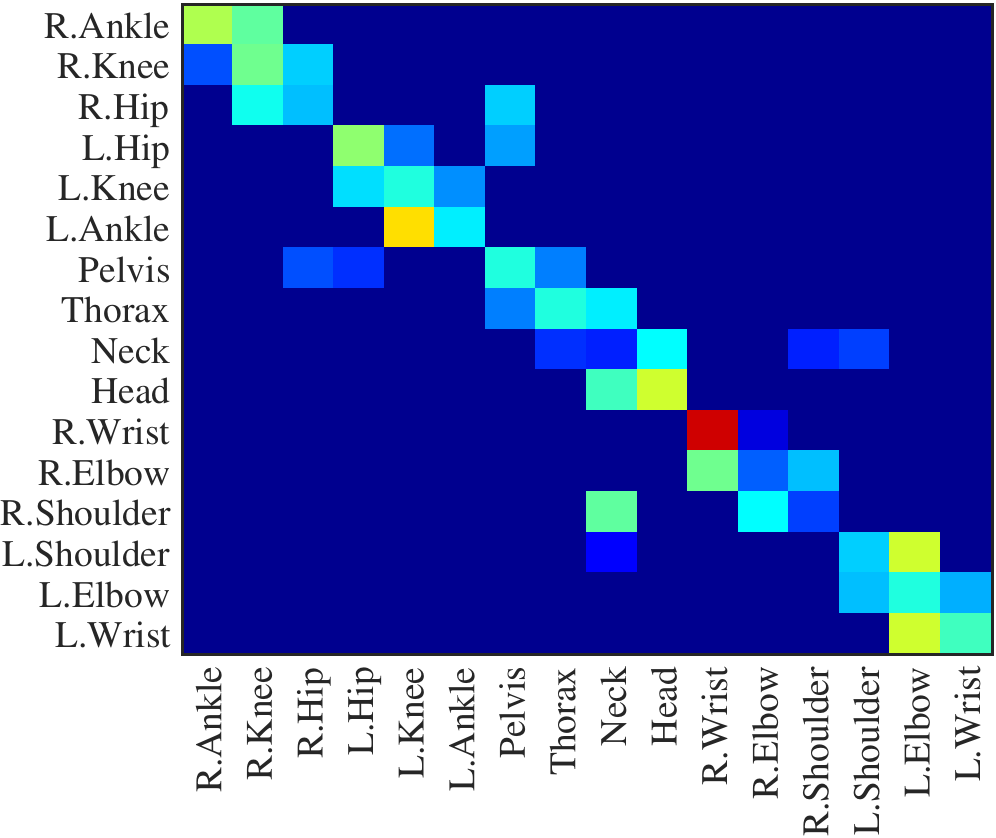}
\includegraphics[height=3.5cm,align=t]{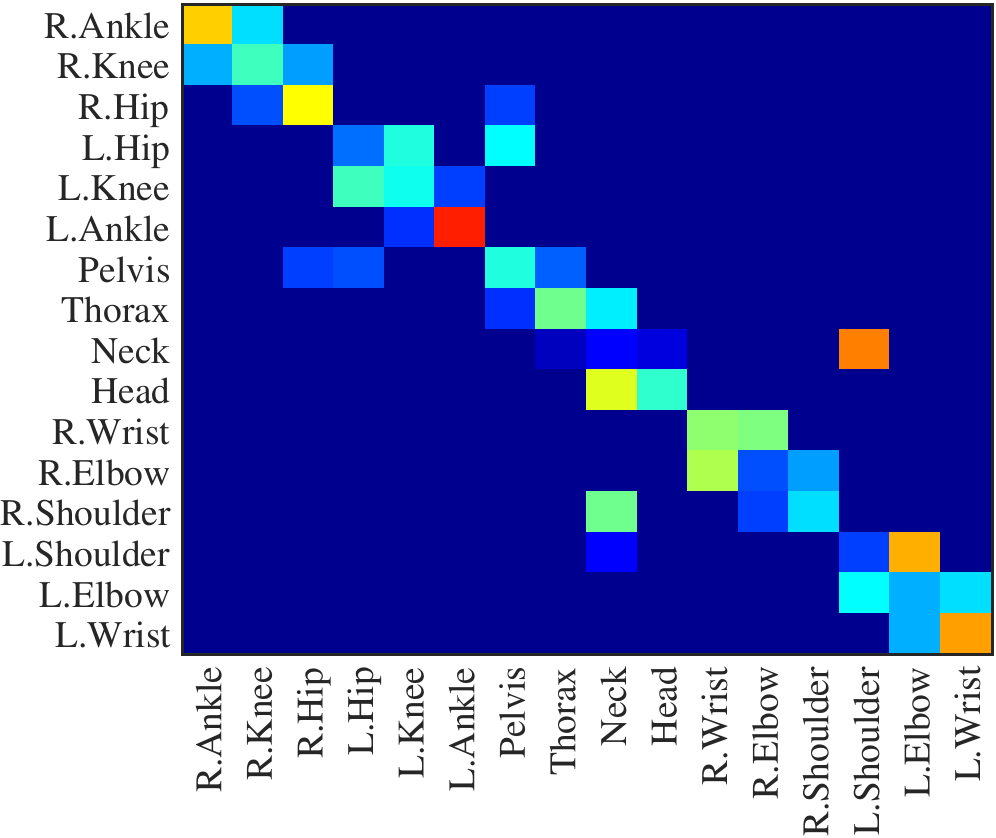}
\includegraphics[height=3.5cm,align=t]{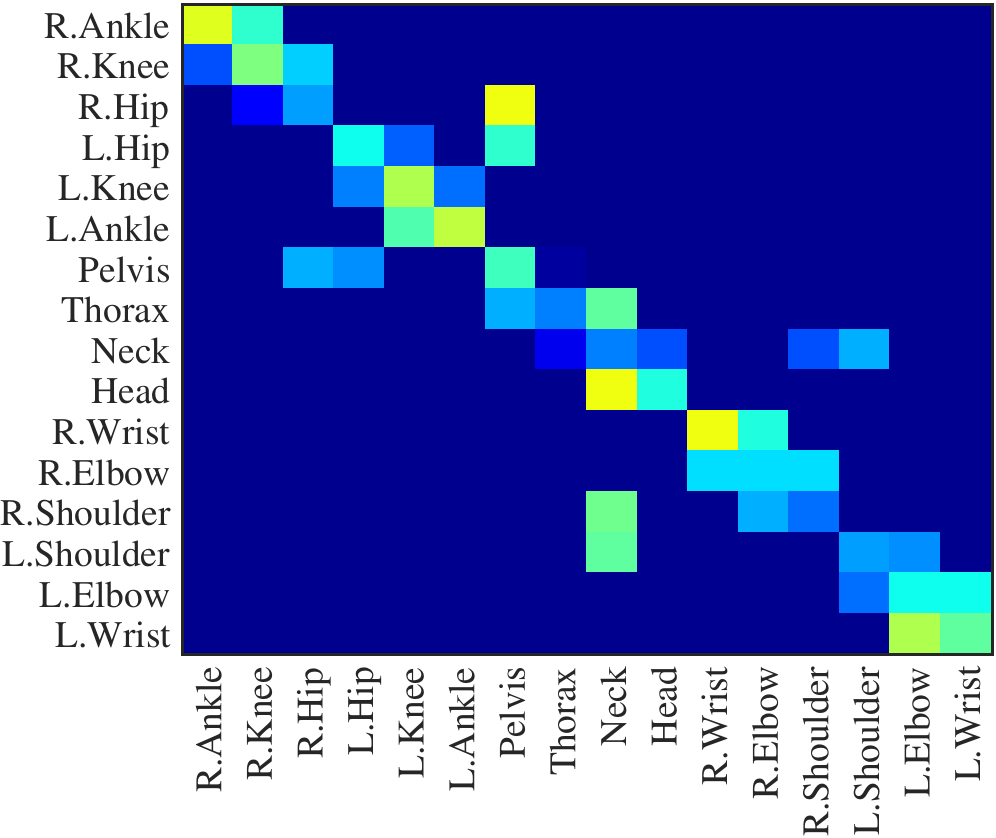}
\includegraphics[height=3.5cm,align=t]{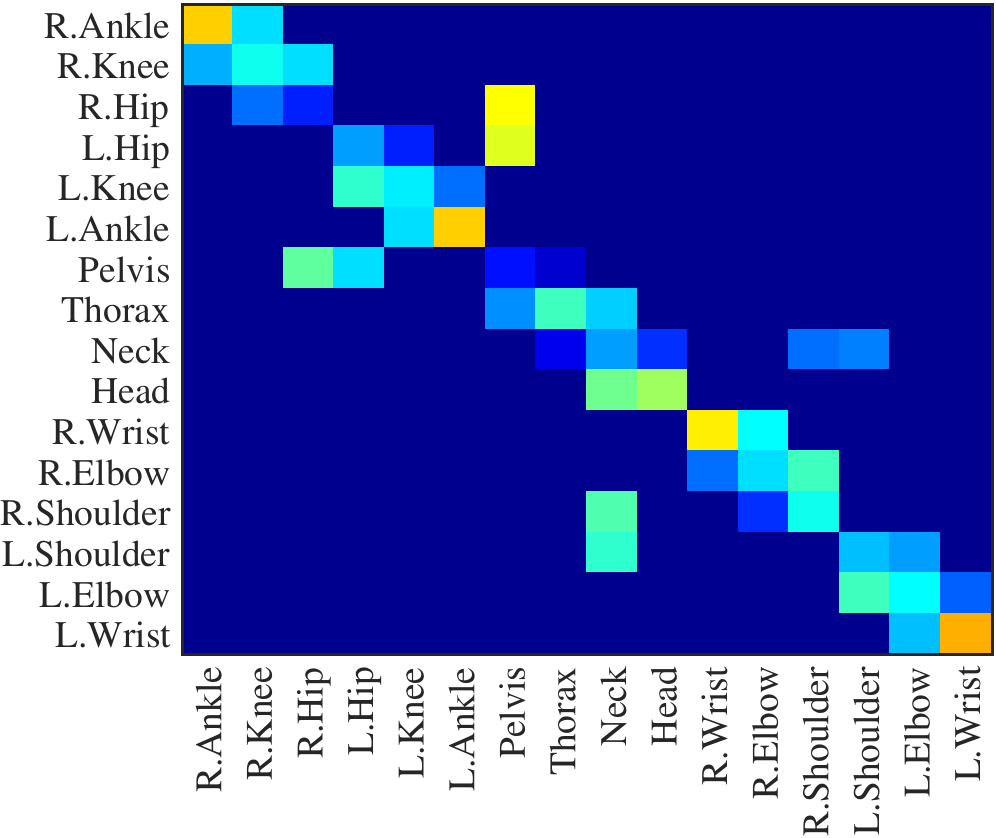}
\includegraphics[height=2.4cm,align=t]{fig70}
\end{center}
   \caption{Visualization of learned weighting matrixes in SemGCN. Each \textit{SemGConv} has learned a different weighting matrix for the graph in the network, while all of them have higher weights for nodes farer from the gravity center of the skeleton. Best viewed in color.}
\label{fig:sgc_l}
\end{figure*}

To better understand the proposed SemGCN, we visualize the learned weighting matrix $\mathbf{M}$ of each \textit{SemGConv} layer in the network. For simplicity, we utilize a simplified version of SemGCN, where Eq.~\ref{eq:ssgc_s} is employed so that all feature channels share the same $\mathbf{M}$. We use the network architecture as illustrated in Fig.~\ref{fig:sgcn_architecture} and train it according to \emph{Configuration \#1} of the experiments.

\begin{figure}
\begin{center}
\includegraphics[width=\linewidth]{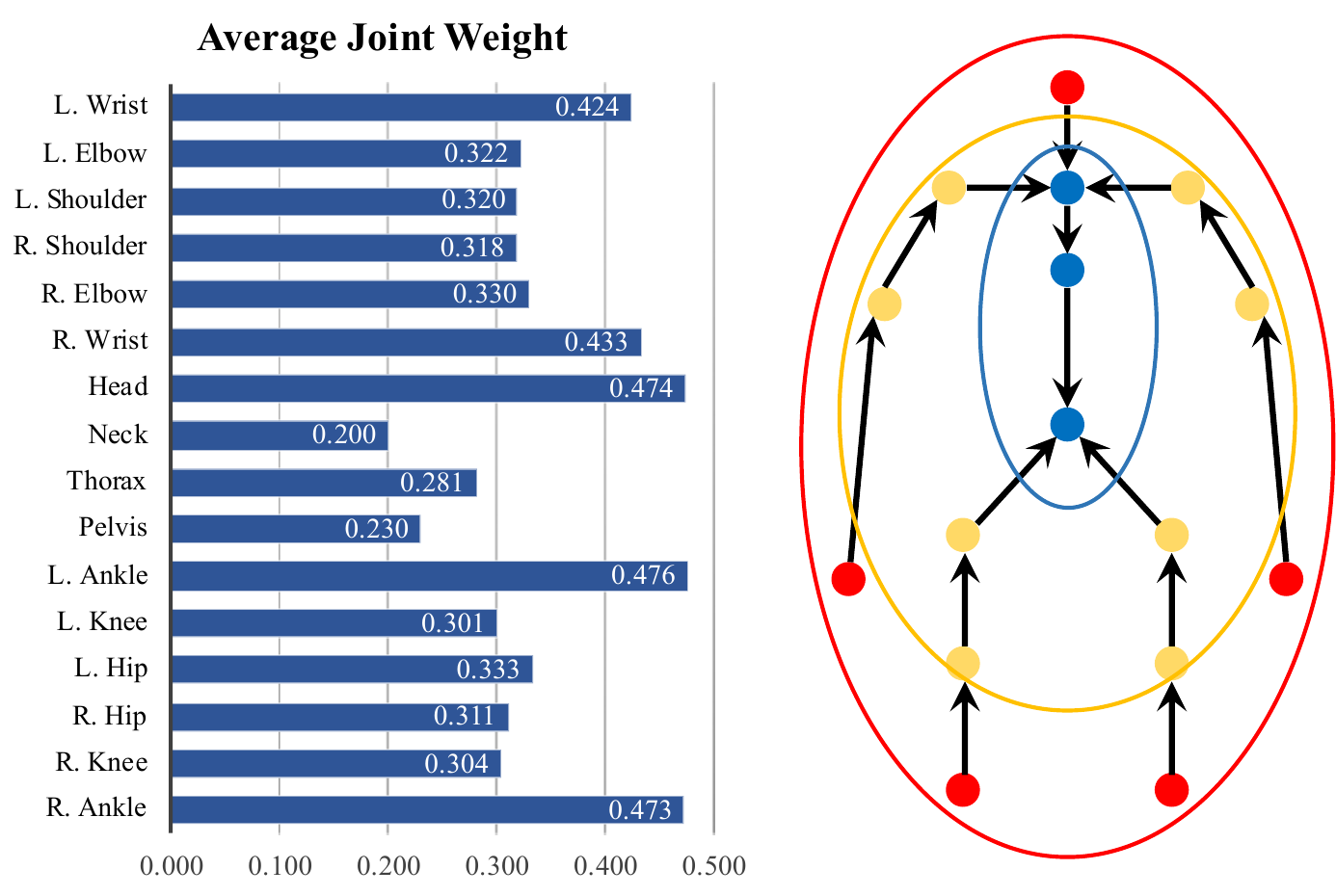}
\end{center}
   \caption{Left: the average learned weight of each joint among all \textit{SemGConv} layers in the network. Right: a regional map of the skeleton where joints are grouped into three regions according to their weights.}
\label{fig:sgc_w}
\end{figure}

The trained network consists of 4 residual blocks where each block contains 2 \textit{SemGConv} layers. Therefore, we visualize the weighting matrixes of these 8 \textit{SemGConv} layers respectively. The matrixes are shown in Fig.~\ref{fig:sgc_l}. We have made two important observations. First, although all \textit{SemGConv} layers share the same graph structure in the network, each of them has learned a different weighting matrix. Second, we can find that \textit{SemGConv} layers have learned higher weights for nodes which are farer from the gravity center of the human skeleton on average.

To further illustrate the second observation, we compute the average learned weight of each joint among all 8 \textit{SemGConv} layers. The quantitative results are shown in Fig.~\ref{fig:sgc_w}(left). We can see that the left wrist, right wrist, left ankle, right ankle and head own the top highest weights which are greater than 0.4; while the neck, thorax and pelvis have the lowest weights less than 0.3. Other joints have quite similar weights around 0.3. This result can be better visualized by representing the human skeleton with a regional map where joints are grouped into three regions according to their weights. Fig.~\ref{fig:sgc_w}(right) shows the result.

This result is intuitive since joints farer from the center always encode more information of the pose while central joints determine the position of the skeleton. This observation is also consistency with~\cite{sun2017compositional,stgcn2018spatial}. This demonstrates that the proposed SemGCN is able to effectively encode spatial relationships of nodes in the graph. However, we only rely on the ground truth for supervision, and no additional hand-crafted constraints or rules are employed.

\end{appendices}

\end{document}